# INVESTIGATING ARTIFICIAL IMMUNE SYSTEMS FOR JOB SHOP RESCHEDULING IN CHANGING ENVIRONMENTS

U Aickelin
E Burke
A Mohamed Din
School of Computer Science & Information Technology
University of Nottingham
Nottingham, UK

## ABSTRACT

Artificial immune system can be used to generate schedules in changing environments and it has been proven to be more robust than schedules developed using a genetic algorithm. Good schedules can be produced especially when the number of the antigens is increased. However, an increase in the range of the antigens had somehow affected the fitness of the immune system. In this research, we are trying to improve the result of the system by rescheduling the same problem using the same method while at the same time maintaining the robustness of the schedules.

## INTRODUCTION

Artificial immune systems are motivated by the theory of immunology. The biological immune system functions to protect the body against pathogens or antigens that could potentially cause harm. It works by producing antibodies that identify, bind to, and finally eliminate the pathogens. Even though the number of antigens is far larger than the number of antibodies, the biological immune system has evolved to allow it to deal with the antigens. The immune system will learn the criteria of the antigens so that in future it can react both to those antigens it has encountered before as well as to entirely new ones. In 2002, de Castro and Timmis suggested that "for a system to be characterized as an artificial immune system, it has to embody at least a basic model of an immune component (e.g. cell, molecule, organ), it has to have been designed using the ideas from theoretical and/or experimental immunology and it has to be aimed at problem solving" [1].

We are dealing with the job shop scheduling problem in this research and we are specifically concerned with tackling the problem of sudden changes in the manufacturing environment. Changes involving unexpected arrival dates of jobs into the factory can result in problems such as jobs to be stored for long periods of time if they arrive early, or cause delays in processing of other jobs if they arrive late. Other possible uncertainties could occur too, such as more new jobs may be introduced or machines may breakdown. Therefore, an efficient scheduling system for the problem should be able to react to such changes as soon as they happen and it should either modify the schedule or generate a new one in order to have an efficient flow in the manufacturing process.

Another important issue in a scheduling problem is the robustness of the schedules generated. According to Sevaux and Sorenson, "a solution is called quality robust if the quality of this solution does not change when the input data changes" [10]. Producing a robust schedule is important especially if we are dealing with a changing environment as building a new schedule might consume a lot of time. Therefore, the schedules generated should, if possible, be able to be used again, may be with a little modification to overcome the problem of uncertainties that occur in the environment. This is the issue that we want to look into in this research.

### An Artificial Immune System Model For Scheduling

Previous work by Hart et al [5], Hart and Ross [6] and Hart and Ross [7], has shown that an artificial immune system model can be used to solve scheduling problems in the manufacturing environment where sudden changes could occur and may need us to produce new schedules even though in the real world situation, such variations in the environment may not be easy to predict. In their research, Hart et al. define the antigen as "the sequence of jobs on a particular machine given a particular scenario" and the antibody as "a short sequence of jobs that is common to more than one schedule" [6].

Using this model, the schedules are generated by first matching the antigen with the antibody. Both the antigen universe and antibody population are generated using a genetic algorithm and are represented by a sequence of integers where the length of the antibodies must be less than the length of the antigens. A genetic algorithm based on GENESIS [4] is used to evolve the antibody where one of the three crossover operators could be used that is, Order-Based Crossover, 2pt-Crossover and Overlap-Crossover. Each antibody is then matched against each of the antigens by aligning an antigen string with an antibody string and calculating a match score.

For example, if there is an antigen string "984567132", which represents the sequence of jobs on one machine, and an antibody string "56789", a partial sequence of jobs from a schedule, we have to align the antibody at every possible alignment position, gene by gene, with the antigen in order to calculate the match score. The match score is calculated by summing up the number of matches where a score of five is given to a perfect match and one is given to a wild card match between the antibody and the antigen. A wild card here is represented by an asterisk as a substitute for any job. Therefore, based on the number of matches between both the antibody and the antigen, the match score for the example given above is 15. The total summation of match scores from the process of matching the antibody with each of the antigens will be assigned to each antibody.

As fitness is important to evolve a diverse range of antibodies, only the antibodies that have the highest match scores will have their match scores added to their fitness functions. These antibodies will then be used as building blocks for building new schedules. Finally, the antibodies are recombined into completed schedules by using three recombination mechanisms: simple recombination, somatic recombination and single job addition. At the same time, the average fitness of the antigens is also calculated to get the fitness of the immune system. This is done from the matching process by calculating the match score where a perfect match is given a score of 0. It is important to know the fitness of the immune system, so that we can see if the immune system is diverse enough to deal with the range of antigens in the antigen universe.

The schedules produced using the model by Hart et al [5] were then compared to the schedules produced by Fang [3] which were developed using a genetic algorithm and the earlier schedules had been shown to be as good as and more robust, in terms of the similarities between each schedule generated, than the latter schedules. However, it was also found that as they increase the number of antigens, the immune system could not give a good performance. Considering this scenario, we are trying to improve the results by rescheduling the same problem using the same method. Fang et al. suggested a proper rescheduling method would be "to re-use some work already done in finding the previous schedule that might involve augmenting the previous schedule with the new change and modifying it until it is acceptable" [2]. Therefore, in our research, the antigen universes will also be generated based on a benchmark problem by Morton & Pentico [9], the same benchmark problem used by Hart et al. and Fang. We will try to increase the number of components within the gene libraries as well as the number of libraries to see if they will affect the performance of the immune system as we increase the number of antigens. A gene library consists of genotypes, which is a collection of genes in the form of a structure, and represented with bit strings [8].

One of the outcomes from Hart et al.'s research is that the immune system can produce good schedules as the range of antigens increases but it affects the performance of the system, in terms of the fitness. This is an indication that the immune system is not diverse enough to cope with the range of antigens. However, as the number of antibodies is increased, the immune system also gives a better performance. This occurs even if the increment is only 0.001% of the set of antibodies constructed [5].

Hart and Ross in [7] pointed out that we could produce a range of antibodies if the components in each library are genetically different. Therefore, we believe that an increment in the number of gene libraries as well as the number of antigens can improve the performance of the immune system. In their research, the antibody molecule is stored in five separate component libraries. Combining a randomly selected component from each library produces a population of 100 artificial immune systems as interpreted by Hart et al. that "an immune system containing $l$ libraries, each with $c$ components, can be used to format $c^l$ different antibodies" [5]. Therefore, in our research, we will generate antibodies using the method by Hightower et al. in [8], which is also used by Hart et al. from up to 10 component libraries and run the same experiments and compare the results with the results produced by them.

The plan for this research is that first, we will evolve the artificial immune system from the gene libraries and generate the antibody population and the antigen universe. Then, we will randomly choose a sample of antigens and antibodies from the antigen universe and the antibody population, respectively. Next is the matching process where the antibody is aligned with each of the antigens and calculates the match scores to get the highest fitness of the antibodies. Finally, we will recombine the antibodies and the partial schedules into a complete schedule.